\documentclass[letterpaper, 10pt, conference]{ieeeconf}  % Comment this line out if you need a4paper

\IEEEoverridecommandlockouts                              % This command is only needed if
\usepackage{amsthm}
\usepackage{times}
\usepackage{multicol}
\usepackage[bookmarks=true]{hyperref}
\usepackage{xcolor}
\usepackage{hyperref}
\usepackage{amsmath, amssymb}
\usepackage{amsfonts}
\usepackage{graphicx}
\usepackage{siunitx}
\usepackage{standalone}
\usepackage{booktabs}
\usepackage[ruled,vlined,linesnumbered,noend]{algorithm2e}
\usepackage{mdframed}
\usepackage{fancyvrb,multirow}
\usepackage{soul}
\usepackage{dsfont,mathabx}
\usepackage{array, booktabs}
\usepackage{subfigure}
\usepackage{booktabs}
\usepackage{makecell}
\usepackage{threeparttable}

\theoremstyle{definition}

\newtheorem{definition}{Definition}

\title{\textbf{BodyGuards: Escorting by Multiple Robots in Unknown Environment\\
under Limited Communication}}

\author{Zhuoli Tian, Yanze Bao and Meng Guo$^*$% <-this % stops a space
  \thanks{$^*$This work was supported by the National Natural Science Foundation of China
    (NSFC) under grants U2241214, T2121002. Corresponding author:
    Meng Guo, meng.guo@pku.edu.cn.}% <-this % stops a space
  }

\begin{document}
\maketitle
\thispagestyle{empty}
\pagestyle{empty}

%%========================================

%%========================================
\begin{abstract}
  Multi-robot systems are increasingly deployed in high-risk missions such as
  reconnaissance, disaster response, and subterranean operations.
  Protecting a human operator while navigating unknown and adversarial
  environments remains a critical challenge,
  especially when the communication among the operator and robots is restricted.
  Unlike existing collaborative exploration methods that aim for complete coverage,
  this work focuses on task-oriented exploration to minimize
  the navigation time of the operator to reach its goal
  while ensuring safety under adversarial threats.
  A novel escorting framework BodyGuards, is proposed
  to explicitly integrate seamlessly collaborative exploration,
  inter-robot-operator communication and escorting.
  The framework consists of three core components:
  (I) a dynamic movement strategy for the operator that maintains a local map
  with risk zones for proactive path planning;
  (II) a dual-mode robotic strategy combining frontier-based exploration
  with optimized return events to balance exploration, threat detection,
  and intermittent communication;
  and (III) multi-robot coordination protocols that jointly plan exploration
  and information sharing for efficient escorting.
  Extensive human-in-the-loop simulations and hardware experiments
  demonstrate that the method significantly reduces operator risk
  and mission time,
  outperforming baselines in adversarial and constrained environments.
\end{abstract}

%%========================================
%==============================
\section{Introduction}\label{sec:intro}

Exploration of unknown and adversarial environments before allowing human access
has become a prominent application for robotic fleets.
Typical scenarios include planetary caves, disaster-stricken areas,
and subterranean tunnels, where safety risks for humans
are high~\cite{klaesson2020planning, petravcek2021large, couceiro2017overview}.
While existing collaborative exploration approaches have achieved impressive results
in mapping and coverage~\cite{burgard2005coordinated, patil2023graph, zhou2023racer},
these methods often prioritize complete workspace coverage.
In contrast, many mission-critical tasks demand \emph{task-oriented exploration},
where the priority is not exhaustive mapping but instead escorting a human operator toward
a target location within an unknown workspace efficiently and safely,
even in the presence of adversarial threats.
In other words,
the operator must be protected from unknown adversaries within unknown environments,
requiring proactive estimation of potential threats in unexplored regions.

At the same time, inter-robot and robot-operator communications are inherently unreliable
in unknown subterranean or obstructed environments, limited to short-range ad-hoc networks.
Exploration and communication are therefore tightly coupled:
advancing exploration without information sharing leaves the operator uninformed,
while too frequent communications hinder task progress.
Designing strategies that balance threat-aware exploration,
operator guidance, and intermittent communication is crucial,
yet remains insufficiently addressed in existing literature.

% ==============================

\begin{figure}[t]
  \centering
  \includegraphics[width=1.0\linewidth]{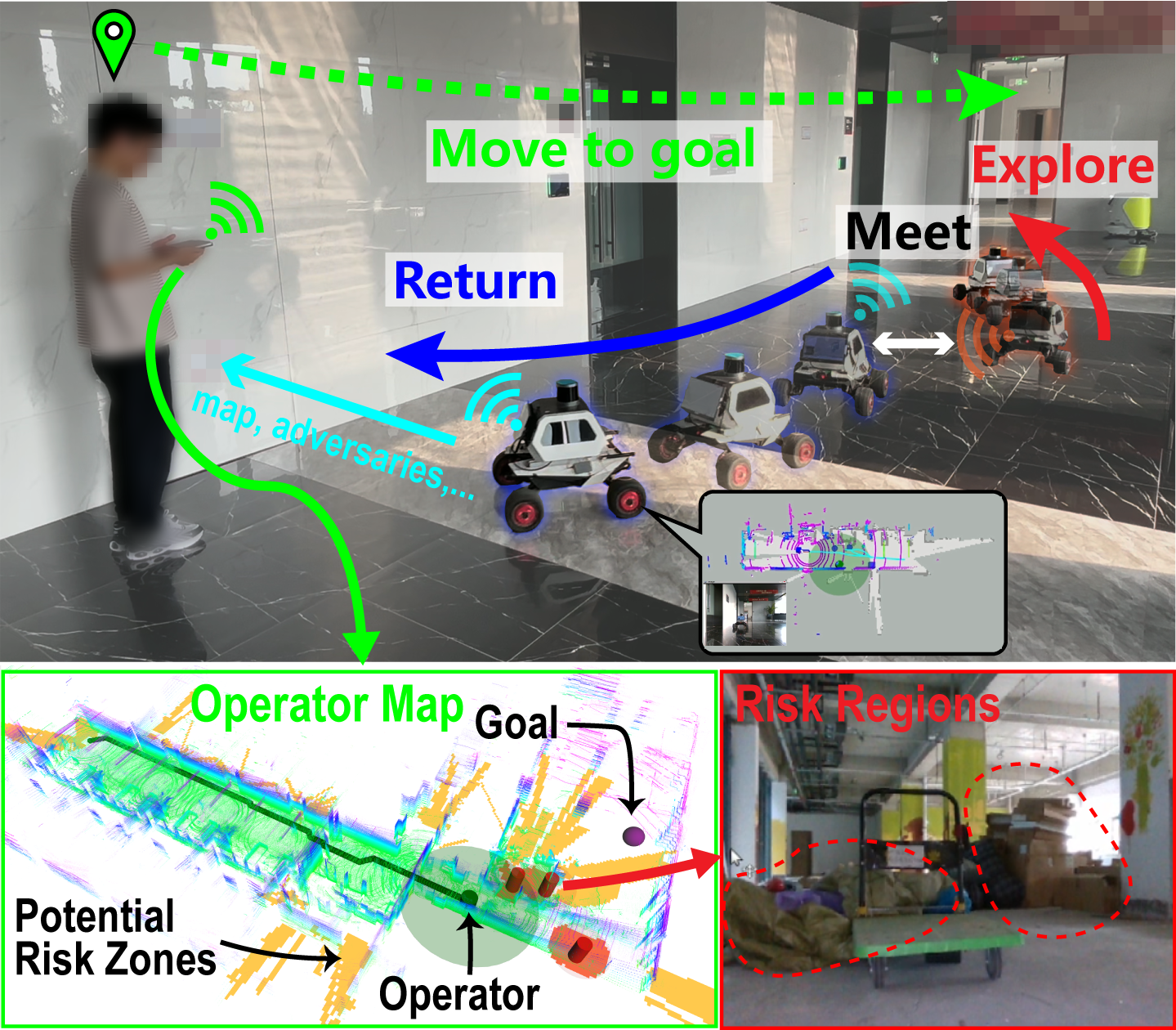}
\vspace{-0.25in}
  \caption{
  \textbf{Top}: Snapshots from hardware experiments showing
      2 UGVs escorting the operator within unknown environment under communication constraints,
  where the robots switch between exploration and communication relay
  to minimize the operator travel time to destination while ensuring safety;
  \textbf{Bottom}: The task-oriented explored map and potential risk regions are provided to the operator
  as guidance.
  }\label{fig:overall}
  \vspace{-0.2in}
\end{figure}
%==============================

%==============================
\subsection{Related Work}
Multi-robot exploration has been extensively studied in the
literature~\cite{burgard2005coordinated, SMMR, patil2023graph, zhou2023racer}.
Most of these approaches focus on collaborative strategies that aim for complete
coverage of the workspace.
Such methods typically assume reliable, all-to-all communication among the robots,
which enables instant sharing of maps and information.
This assumption is impractical in subterranean or obstructed environments, where
inter-robot communication is constrained by range and bandwidth limitations~\cite{esposito2006maintaining, tian2024ihero}.
To address this, several works have incorporated planning for both communication
and exploration, including intermittent relay-based
strategies~\cite{gao2022meeting, guo2018multirobot, saboia2022achord}.
However, these methods still emphasize coverage and information gain, rather than
task-oriented objectives.
In contrast, task or goal-oriented exploration has recently emerged,
where the priority is to minimize the mission completion time for specific objectives
such as reaching a goal location and collecting feature information~\cite{9645287, 10517451}.
This paradigm shift is particularly relevant when communication must be planned
together with exploration and task requirements~\cite{flykites}.
However, most of these works neglect the online interaction with human operators.

Indeed, the role of the human operator further distinguishes task-oriented escorting
from coverage-driven exploration.
While robotic fleets are increasingly autonomous, the operator remains essential
for decision-making, confirmation, and direct intervention during execution~\cite{dahiya2023survey}.
Prior works on human-fleet coordination have explored teleoperation,
mixed autonomy, and augmented-reality interfaces to improve situational
awareness~\cite{reardon2019communicating, gao2022meeting}.
Nevertheless, these efforts often assume global communication links and do not address
the escorting problem explicitly.
Escorting requires not only guiding the operator safely through unknown and adversarial
environments, but also ensuring timely information update despite limited connectivity,
to avoid leaving the operator uninformed or unprotected in critical stages~\cite{2021intention}.
Existing studies on operator-robot interaction~\cite{riley2023fielded} and multi-robot escorting
strategies~\cite{2023escort} often requires a known workspace or perfect communication.
Thus, a unified framework that couples task-oriented exploration,
intermittent communication, and human escorting remains largely unexplored.

%==============================
\subsection{Our Method}
To address these challenges, this work proposes \textbf{BodyGuards},
a framework for escorting an operator in adversarial, unknown, and communication-constrained environments.
The method integrates risk estimation, dynamic path planning, and multi-robot-operator coordination into a unified structure.
At its core, the operator's path is guided by a continuously updated map with risk zones that anticipate potential adversaries at unexplored frontiers,
which prevents the operator from entering unsafe regions.
In parallel, the robotic fleet alternates between frontier-based exploration and optimized return events to relay information,
of which the key is to estimate how much the newly explored area might contribute
to the dynamic shortest path of the operator.
These activities are jointly optimized to minimize operator arrival time.
Extensive human-in-the-loop simulations and hardware experiments demonstrate that the method significantly reduces operator risk and mission time compared to state-of-the-art baselines.

The contributions of this work are twofold:
(I) a novel formulation of task-oriented exploration is introduced for escorting
an operator in adversarial and unknown environments under limited communication;
and (II) an integrated framework combining the risk-aware operator guidance, collaborative
task-oriented exploration,
and coordination via intermittent communication is developed.

%%========================================
\section{Problem Description}\label{sec:problem}

%==============================
\subsection{Robots and Operator in Workspace}\label{subsec:ws}
Consider a 2D workspace~$\mathcal{W}\subset \mathbb{R}^2$,
of which its {map} including the boundary, freespace
and obstacles are all \emph{unknown}.
A team of robots denoted by~$\mathcal{N}\triangleq\{1,\cdots,N\}$
is deployed by an operator to explore the workspace.
Each robot~$i\in \mathcal{N}$ is capable of simultaneous localization
and mapping (SLAM)~\cite{tian2022kimera}
and has a navigation module with a maximum speed~$v_{r}>0$.
Denote by~$p_i(t)\in \mathcal{W}$ the 2D pose and $\mathcal{M}_i(t)\subseteq \mathcal{W}$
the local map of robot~$i$ at time~$t>0$ that contains the known free space.
Analogously, the operator has a~2D pose~$p_{\texttt{h}}(t)\in \mathcal{W}$
and a local map~$\mathcal{M}_\texttt{h}(t)\subseteq \mathcal{W}$,
along with the mobility of a maximum speed~$v_{\texttt{h}}>0$.
For brevity, denote by~$\mathcal{N}^+\triangleq \mathcal{N}\cup \{\texttt{h}\}$.

% ==============================
\begin{figure}[t!]
  \centering
  \includegraphics[width=1.0\linewidth]{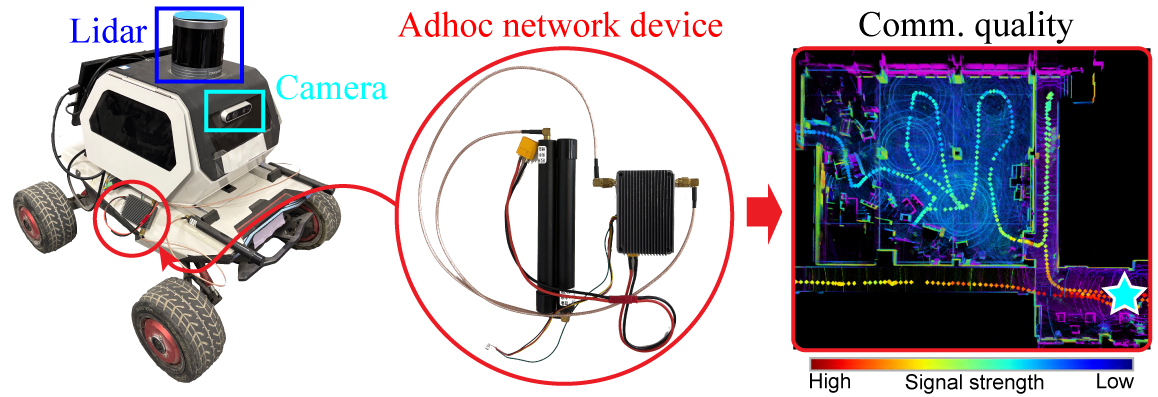}
  \vspace{-0.3in}
  \caption{
  Each robot and the operator is equipped with a communication module
  to exchange information with each other,
  and the signal strength changes as they move within the workspace.
  % \todo{add the operator with a tablet.}
  }\label{fig:adhoc_device}
  \vspace{-0.2in}
\end{figure}
%==============================

Moreover, as illustrated in Fig.~\ref{fig:adhoc_device},
the robots and the operator are equipped
with a communication module to exchange data.
Namely, each robot~$i\in \mathcal{N}$
can exchange information (including the local maps)
with another robot~$j\in \mathcal{N}^+$
or the operator via wireless communication,
if the communication quality between them is above a threshold,
%==============================
\begin{equation}\label{eq:com}
  \texttt{Com}_{ij}(p_i,p_j,\mathcal{M}_i)>\underline{c};\,
  \mathcal{M}_{\texttt{h}}(t)\subseteq \bigcup_{i\in \mathcal{N}} \mathcal{M}_i(t),
\end{equation}
%==============================
where~$\underline{c}>0$ is the minimum signal strength required
for successful communication in the workspace;
and the operator can only obtain local maps from the robots.
Thus, the behavior of each robot~$i\in \mathcal{N}$ is determined
by its timed sequence of navigation and communication events, i.e.,
%==============================
\begin{equation}\label{eq:behavior}
\Gamma_i\triangleq c^0_i\, \mathbf{p}^0_i\, c^1_i \, \mathbf{p}^1_i\,c^2_i \, \cdots,
\end{equation}
%==============================
where~$c^m_i\triangleq (j,\, p_{ij},\, t_{ij})$ is
the communication event with robot~$j\in \mathcal{N}_i(t)$
at location~$p_{ij}\in \mathcal{W}$ at time~$t_{ij}$;
the navigation path~$\mathbf{p}^m_i\subset \mathcal{W}$ contains the path
between these communication events.
Similarly, the behavior of the operator is denoted by
$\Gamma_{\texttt{h}}\triangleq c^0_{\texttt{h}}\, \mathbf{p}^0_{\texttt{h}}\, c^1_{\texttt{h}} \,
\mathbf{p}^1_{\texttt{h}}\,c^2_{\texttt{h}} \, \cdots$.
For brevity, denote by $\widehat{\Gamma}(t)\triangleq (\{\Gamma_i(t)\}, \Gamma_{\texttt{h}}(t))$ the joint
behaviors of the robots and the operator by time $t\geq 0$.

%==============================
\subsection{Escorting Task with Adversarial Regions}\label{subsec:escort}
Within the workspace~$\mathcal{W}$,
there are in total~$M>0$ \emph{static} adversaries, denoted by~$\mathcal{A}\triangleq\{1,\cdots,M\}$.
Each adversary~$m\in \mathcal{A}$ has an \emph{unknown} position~$a_m \in \mathcal{W}$,
and a \emph{known} risk range~$r_{m}>0$ that encloses a risk region.
%=====================
\begin{definition}[Risk Region]\label{def:dangerous-region}
  The risk region associated with adversary~$m\in \mathcal{A}$ is defined as
  the area that has a line-of-sight within the risk range, i.e.,
  $\mathcal{D}_m\triangleq\{p\in \texttt{LOS}(a_m,\, \mathcal{W})\mid
  \|p-a_m\| \leq r_m\}$,
  where~$\texttt{LOS}(a_m,\, \mathcal{W})$ contains the visible area
  from position~$a_m$ within~$\mathcal{W}$.
  \hfill $\blacksquare$
\end{definition}
%=====================

The risk region of all adversaries are all initially unknown,
which however can be detected by the robots with a line-of-sight and
within their sensor range~$r_\texttt{s}>0$.
Initially, the operator and robots are deployed at the same location~$p_\texttt{S}$,
and their overall mission is to escort the operator to a target location~$p_\texttt{G}\in \mathcal{W}$,
which is initially within the unknown workspace.
To ensure safety,
the operator should:
(I) move within the known obstacle-free area of its local map~$\mathcal{M}_{\texttt{h}}$;
and (II) avoid all risk regions of all adversaries, i.e.,
%==============================
\begin{equation}\label{eq:safety}
p_{\texttt{h}}(t) \in \mathcal{M}_{\texttt{h}}(t) \setminus \{\mathcal{D}_m,\forall m\in \mathcal{A}\},
\end{equation}
%==============================
both of which can only be known to the operator online
when a robot actively returns to communicate, due to the limited communication.
The escorting task is considered completed when the operator reaches the target location safely.

%==============================
\subsection{Problem Statement}\label{subsec:problem}
The overall problem is formalized as a constrained optimization over
the collaborative exploration and communication strategy over the fleet and the operator,
i.e.,
%====================
\begin{equation} \label{eq:problem}
  \begin{aligned}
    &\mathop{\mathbf{min}}\limits_{\{\widehat{\Gamma},\, \overline{T}\}}\;  \overline{T}\\
    \textbf{s.t.}\quad & p_\texttt{h}(\overline{T}) = p_\texttt{G};\,
    \eqref{eq:com}-\eqref{eq:safety},\,
    \forall t\leq\overline{T};
  \end{aligned}
\end{equation}
%====================
where~$\overline{T}>0$ is the total time when the operator
reaches the target location~$p_\texttt{G}$ subject to the aforementioned
communication and safety constraints.

%%========================================
 \section{Proposed Solution}\label{sec:solution}
 The BodyGuards framework integrates the following components.
 First, the dynamic operator movement is guided by the risk-aware shortest path
 in the operator local map
 to proactively avoid adversarial exposure, as described in Sec.~\ref{subsec:operator}.
 Then, the robot strategy is presented in Sec.~\ref{subsec:single}
 such that it alternate between exploration and communication
 to balance exploration, detection of adversaries, and return events to the operator.
 Furthermore, the protocol for inter-robot intermittent communication
 is described in Sec.~\ref{subsec:multi},
 to enable collaborative exploration and more timely update of the operator map.
 Lastly, the strategy for online execution and generalizations
 are presented in Sec.~\ref{subsec:online}.

%%========================================
%==============================================
\subsection{Dynamic Motion Strategy for the Operator}
\label{subsec:operator}

The operator often cannot, in general, reach the global goal~$p_{\texttt{G}}$
directly due to unknown workspace and risk regions.
Thus, it follows the motion strategy that dynamically selects a temporary goal~$p_{\texttt{g}}$
within its local map~$\mathcal{M}_{\texttt{h}}$,
and then follows the shortest and safe path towards~$p_{\texttt{g}}$.
It involves the following three sequential steps.

%==============================
\subsubsection{Potential Risk Zones}
\label{subsubsec:TEZ}
\begin{figure}[t!]
  \centering
  \includegraphics[width=1.0\linewidth]{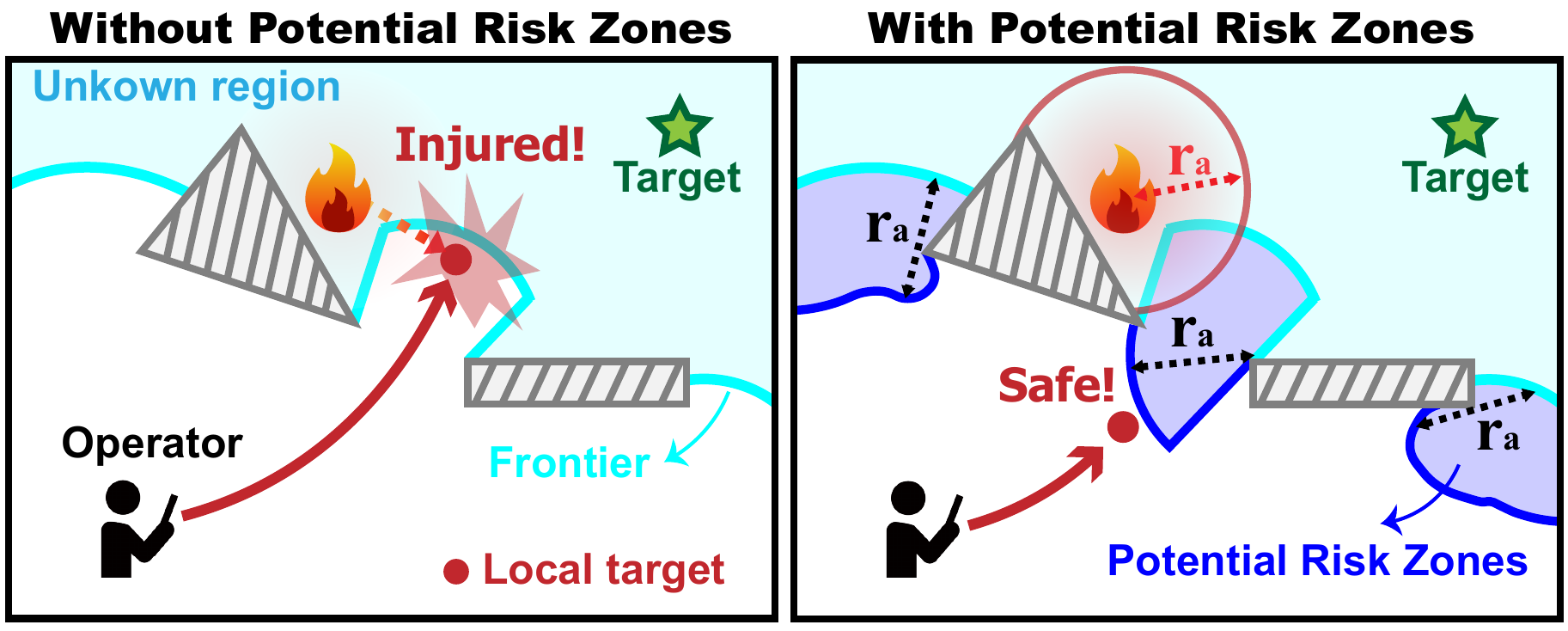}
  \vspace{-0.3in}
  \caption{
  \textbf{Left}: With potential adversaries within unknown space,
  the operator may have risks when reaching the frontiers;
  \textbf{Right}: when the proposed Potential Risk Zones (in blue) is enforced,
  the operator remains safe.
  }\label{fig:ZTZ_illu}
  \vspace{-0.2in}
\end{figure}

Since the operator local map~$\mathcal{M}_{\texttt{h}}$ is only partial,
many areas on its boundary could belong to the risk region of potential adversaries
in unknown parts.
As illustrated in Fig.~\ref{fig:ZTZ_illu},
it is important to exclude such areas from the set of temporary goals for safe navigation.

%=====================
\begin{definition}[Potential Risk Zones]\label{def:TEZ}
  Given the local map of the operator~$\mathcal{M}_\texttt{h}(t)$,
  the potential risk zones are defined as the area that could belong to a risk region
  if there are potential adversaries close
  to its boundary~$\partial \mathcal{M}_\texttt{h}(t)$ excluding the workspace boundary,
  i.e.,
  \begin{equation}\label{eq:TEZ}
    \mathcal{Z}_\texttt{h}(t) \triangleq \bigcup_{p \in \partial \mathcal{M}_\texttt{h}(t)\setminus \partial \mathcal{W}}\,
    \Upsilon\big{(}p,\,\mathcal{M}_\texttt{h}(t)\big{)},
  \end{equation}
  where $\Upsilon(p,\,\mathcal{M}_\texttt{h}(t))\subset \mathcal{M}_\texttt{h}(t)$
  is the risk region defined similarly to Def.~\ref{def:dangerous-region}.
  \hfill $\blacksquare$
\end{definition}
% =====================
The potential risk zones~$\mathcal{Z}_\texttt{h}(t)$ can be computed
by sufficiently sampling the boundary points and computing the union
of their risk regions, as shown in Fig.~\ref{fig:ZTZ_illu}.

\subsubsection{Dynamic Update of Operator Map for Safety}
\label{subsubsec:dynamic-map-refinement}
Given $\mathcal{Z}_\texttt{h}(t)$ and the set of known adversaries,
the operator map is updated as follows for safety.
First, denote by~$\mathcal{A}_\texttt{h}(t)\triangleq \{{i_1}, {i_2}, \cdots, {i_K}\}\subseteq \mathcal{A}$
the set of adversaries known to the operator at time~$t>0$.
Consequently, the associated risk regions are given by
$\mathcal{D}_\texttt{h}(t) \triangleq \bigcup_{\ell=1}^K \mathcal{D}_{i_\ell}$,
where~$\mathcal{D}_{i_\ell}$ is derived by Def.~\ref{def:dangerous-region} for adversary~${i_\ell}$.
Then, the local map of the operator is updated as follows:
\begin{equation}\label{eq:map-update}
  \overline{\mathcal{M}}_\texttt{h}(t) \triangleq
  \texttt{UpdateSafe}(\mathcal{M}_\texttt{h}(t), \mathcal{Z}_\texttt{h}(t), \mathcal{D}_\texttt{h}(t)),
\end{equation}
where function~$\texttt{UpdateSafe}(\cdot)$
marks the area~$\mathcal{Z}_\texttt{h}(t)$ as unknown,
and the area~$\mathcal{D}_\texttt{h}(t)$ as unsafe.
Thus, it is only safe for the operator to stay outside of both~$\mathcal{Z}_\texttt{h}(t)$ and $\mathcal{D}_\texttt{h}(t)$.

%=======================
\subsubsection{Choice of Temporary Goal}
\label{subsubsec:local-target-optimization}
\begin{figure}[t]
  \centering
  \includegraphics[width=1.0\linewidth]{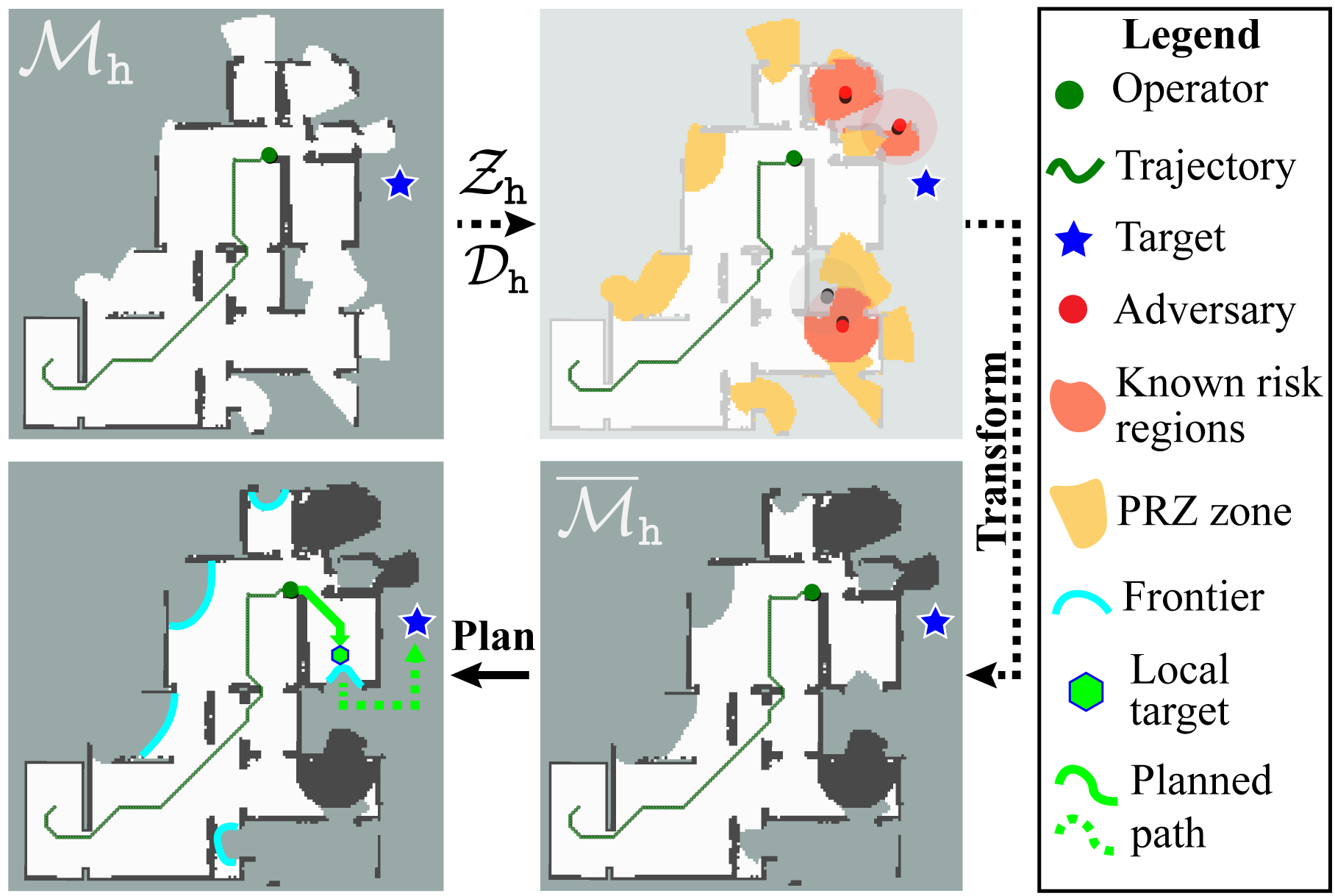}
  \vspace{-0.3in}
  \caption{
  The operator dynamically refines its local map by excluding the potential risk zones~$\mathcal{Z}_\texttt{h}(t)$
  and the known risk regions~$\mathcal{D}_\texttt{h}(t)$,
  and then selects a temporary goal~$\widehat{p}_\texttt{g}(t)$ on the refined map to move towards.
  }\label{fig:refine_map}
  \vspace{-0.2in}
\end{figure}

As shown in Fig.~\ref{fig:refine_map}, the temporary goal is chosen on the boundary of updated map~$\overline{\mathcal{M}}_\texttt{h}(t)$.
It minimizes the estimated distance that the operator needs to travel to reach the target~$p_\texttt{G}$.

 % ==============================
%==============================

%=====================
\begin{definition}[Estimated-Via-Distance]\label{def:via-distance}
  Given three points~$p_1,p_2,p_3$ within a partially-known map~$\mathcal{M}$,
  the estimated distance from~$p_1$ to~$p_3$ via~$p_2$ is defined as:
  \begin{equation}\label{eq:via-distance}
    \chi(p_1, p_2, p_3, \mathcal{M}) \triangleq d(p_1, p_2, \mathcal{M}) + g(p_2, p_3, \mathcal{M}),
  \end{equation}
  %==============================
  where~$d(\cdot)$ is the~$\textup{A}^\star$ distance from $p_1$ to $p_2$
  within the known part of~$\mathcal{M}$;
  $g(\cdot)$ is the \emph{estimated}~$\textup{A}^\star$
  distance from~$p_2$ to~$p_3$ by treating the unknown part of~$\mathcal{M}$ as obstacle-free.
  \hfill $\blacksquare$
\end{definition}
%=====================
Denote by $\overline{\mathcal{F}}_\texttt{h}(t)\triangleq \{\overline{f}_\texttt{h}^k\}$ its set of frontiers~\cite{holz2010evaluating}
on the boundary~$\partial \overline{\mathcal{M}}_\texttt{h}(t)$ between the known and unknown areas,
which can be identified via a Breadth-First-Search (BFS) and then clustered by various techniques.
Consequently, the temporary goal~$\widehat{p}_\texttt{g}(t)$ for the operator can be selected by:
%==============================
\begin{equation}\label{eq:temporary-h-goal}
  \widehat{p}_\texttt{g}(t) \triangleq \underset{{\overline{f}_\texttt{h}^k\in
      \overline{\mathcal{F}}_\texttt{h}(t)}}{\textbf{argmin}}\,
  \big{\{}\chi\big(p_\texttt{h}(t),\, \overline{f}_\texttt{h}^k,\,
  p_\texttt{G},\, \overline{\mathcal{M}}_\texttt{h}(t)\big)\big{\}},
\end{equation}
%==============================
where the estimated-via-distance treats the potential risk zones~$\mathcal{Z}_\texttt{h}(t)$ unknown
and obstacle-free by Def.~\ref{def:via-distance}.
In other words, the temporary goal minimizes the estimated distance that the operator
needs to travel to reach the target~$p_\texttt{G}$ given its current updated map~$\overline{\mathcal{M}}_\texttt{h}(t)$.
Once the temporary goal~$\widehat{p}_\texttt{g}(t)$ is selected,
the operator then move towards it following the shortest path with its maximum speed~$v_{\texttt{h}}$,
i.e.,
%==============================
\begin{equation}\label{eq:operator-plan}
  \Gamma_\texttt{h}(t)\triangleq \texttt{SafePath}
  \big(p_\texttt{h}(t),\, \widehat{p}_\texttt{g}(t),\, \overline{\mathcal{M}}_\texttt{h}(t)\big),
\end{equation}
%==============================
where~$\texttt{SafePath}(\cdot)$ returns the~$\textup{A}^\star$ shortest safe path
between two points within the known and safe part of the map.
% ==============================
%==============================

%%========================================
%%========================================
%==============================================
\subsection{Dual-mode Robot Strategy}
\label{subsec:single}

As described in~\eqref{eq:map-update},
it is utterly important for the robot to explore the workspace,
and detect obstacles and adversaries therein.
More importantly, these information should be shared with the operator
via local communication.
This section describes the robot strategy to switch
between exploration and return events to the operator for communication.

\subsubsection{Frontier-based Exploration}
Let $\mathcal{M}_\texttt{r}(t)$ be the local map of any robot~$i\in \mathcal{N}$ at time~$t>0$,
and $\mathcal{F}_\texttt{r}(t)\triangleq {f_\texttt{r}^k}$ the set of frontiers between explored and unexplored areas, serving as exploration targets.
Then, the cost of each frontier~$f_\texttt{r}^k\in \mathcal{F}_\texttt{r}(t)$ is given by:
%==============================
\begin{equation} \label{eq:next-frontier}
  \begin{aligned}
   \eta(f_\texttt{r}^k,\,\mathcal{M}_\texttt{r}) \triangleq\
     & w_1\, d(p_\texttt{r},\, f_\texttt{r}^k,\, \mathcal{M}_\texttt{r})\\
     & + w_2\, \chi(\overline{p}_\texttt{h},\,
     f_\texttt{r}^k,\, p_\texttt{G},\, \overline{\mathcal{M}}_\texttt{r}),
\end{aligned}
\end{equation}
%==============================
where $w_1,w_2>0$ are weights, the first term is the robot's travel distance to the frontier,
and the second term is the estimated distance for the operator to reach its target if the area near $f_\texttt{r}^k$ is explored.
Here, $\overline{p}_\texttt{h}$ is the estimated human position,
and $\overline{\mathcal{M}}_\texttt{r}$ is the operator's estimated local map from updating $\mathcal{M}_\texttt{r}$ via~\eqref{eq:map-update}.
The frontier with minimum cost $\widehat{f}_\texttt{r}$ is selected,
and the robot navigates to it via the A* path in $\mathcal{M}_\texttt{r}(t)$,
yielding local plan $\Gamma_\texttt{r}$.

%==============================
\subsubsection{Optimization of Return Events}
More importantly,
the robot should also optimize the return events to communicate with the operator
intermittently to update its local map~$\mathcal{M}_\texttt{h}(t)$ and adversary information~$\mathcal{A}_\texttt{h}(t)$.
Since the operator has no knowledge of this meeting event due to limited communication,
the meeting location should be selected only on the planned path of
operator~$\Gamma_\texttt{h}(t)$ by~\eqref{eq:operator-plan} within a similar time window.
The objective is to minimize the estimated time for the operator to reach the target~$p_\texttt{G}$,
if the operator is informed of the latest information at location~$p_\texttt{com}$ and time~$t_\texttt{com}$.
Denote by~$\mathcal{M}'_\texttt{h}$ the local map of the operator \emph{after} the communication event,
$\overline{\mathcal{M}}'_\texttt{h}$ the updated map by~\eqref{eq:map-update},
and~$\overline{\mathcal{F}}'_\texttt{h}$ the associated frontiers.
 % ==============================
\begin{figure}[t!]
  \centering
  \includegraphics[width=1.0\linewidth]{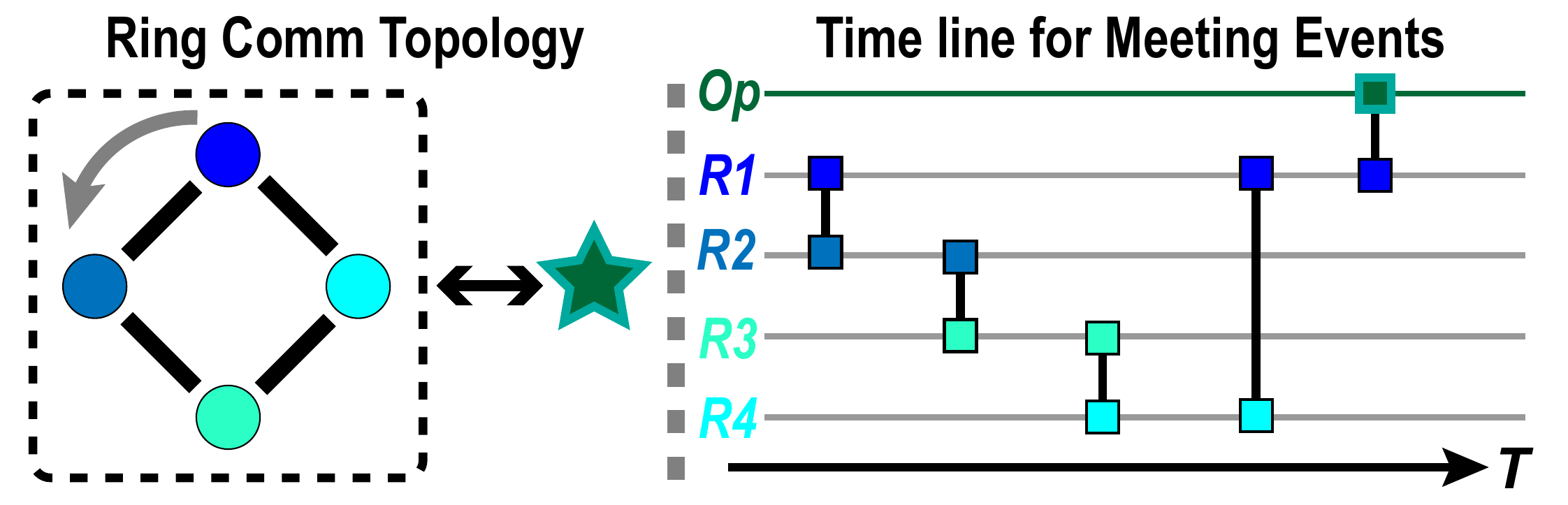}
  \vspace{-0.3in}
  \caption{
    \textbf{Left}:
    The proposed ring communication topology among multiple robots
    (filled circles) and the operator (filled star);
  \textbf{Right}: the communication events along the time axis.
  }\label{fig:Ring_topo}
  \vspace{-0.2in}
\end{figure}
%==============================
Then, the subproblem of optimizing the return event is formulated as:
%====================
\begin{subequations} \label{eq:sub-problem}
  \begin{align}
    \mathop{\mathbf{min}}\limits_{p'_\texttt{r}, p_\texttt{com}, t_\texttt{com}, t_\texttt{w}}\;
    &\Big{\{}t_\texttt{com}+ \frac{\chi(p_\texttt{com},\, p'_\texttt{g},\, p_\texttt{G},\,
      \overline{\mathcal{M}}'_\texttt{h})}{v_\texttt{h}}\Big{\}} \notag\\
    \textbf{s.t.}\quad
    &t_\texttt{com}= t+\frac{d(p_\texttt{h},\, p_\texttt{com},\, \overline{\mathcal{M}}'_\texttt{h})}{v_\texttt{h}}
    +t_\texttt{w}; \label{subeq:t_com-1}\\
    &t_\texttt{com}\geq t+\frac{\chi(p_\texttt{r},\, p'_\texttt{r},\, p_\texttt{com},\, \mathcal{M}_\texttt{r})}{v_\texttt{r}}; \label{subeq:t_com-2}\\
    &t_\texttt{w}\geq 0; \label{subeq:wait-time}\\
    &t_\texttt{w}\,\| p_\texttt{com}-\widehat{p}_\texttt{g}\|=0; \label{subeq:com-time}\\
    &p'_\texttt{g}=\underset{f\in \overline{\mathcal{F}}'_\texttt{h}}{\textbf{argmin}}\,
    \big{\{}\chi(p_\texttt{com},\, f,\, p_\texttt{G},\, \overline{\mathcal{M}}'_\texttt{h})\big{\}}; \label{subeq:p_g}
  \end{align}
\end{subequations}
%====================,
where~$p'_\texttt{r}\in {\Gamma}_\texttt{r}$ is a waypoint within the robot plan before return;
$p_\texttt{com}\in {\Gamma}_\texttt{h}$ is the communication location within the operator plan;
$t_\texttt{w}\geq 0$ is the waiting time for the operator after arriving at the communication location~$p_\texttt{com}$.
The first three constraints in~\eqref{subeq:t_com-1}-\eqref{subeq:wait-time} ensure that
the communication event happens only after the robot and operator reach their respective locations;
the constraint~\eqref{subeq:com-time} ensures that the operator \emph{can only wait}
at the current temporary goal~$\widehat{p}_\texttt{g}$ as the operator has no knowledge of the communication event;
and the last condition in~\eqref{subeq:p_g} computes the next optimal temporary goal \emph{after} the communication
by~\eqref{eq:temporary-h-goal}.
This problem can be solved by first searching over all choices of
$p'_\texttt{r}\in {\Gamma}_\texttt{r}$ and $p_\texttt{com}\in {\Gamma}_\texttt{h}$,
and then formulating a linear program (LP) over~$t_{\texttt{com}}$ and $t_{\texttt{w}}$,
which can be solved by a commercial solver~\cite{gor}.
Given the solution of~\eqref{eq:sub-problem}, the complete hybrid plan of the robot is given by:
\begin{equation}\label{eq:robot-plan}
    \widehat{\Gamma}_\texttt{r}(t) \triangleq \texttt{HybPlan}(p_\texttt{r},\, p'_\texttt{r},\, p_\texttt{com},\, \mathcal{M}_\texttt{r}),
\end{equation}
where the robot first follows the old path~$\Gamma_{\texttt{r}}$ to reach~$p'_\texttt{r}$;
then follows the~$A^\star$ shortest path to the communication location~$p_\texttt{com}$;
and finally the communication event to exchange information with the operator around time~$t_{\texttt{com}}$,
including its local map~$\mathcal{M}_\texttt{r}(t)$, adversaries~$\mathcal{A}_{\texttt{r}}(t)$
and the temporary goal~$\widehat{p}_{\texttt{g}}$ and updated operator map~$\overline{\mathcal{M}}_\texttt{h}(t)$.

%%========================================
%%========================================
%==============================================
\subsection{Multi-robot Coordination via Intermit. Communication}
\label{subsec:multi}
Efficiency of the above escorting strategy can be further improved
by employing multiple robots.
This section introduces the multi-robot coordination strategy
via the intermittent communication protocol.

% ==============================
\begin{figure}[t!]
  \centering
  \includegraphics[width=1.0\linewidth]{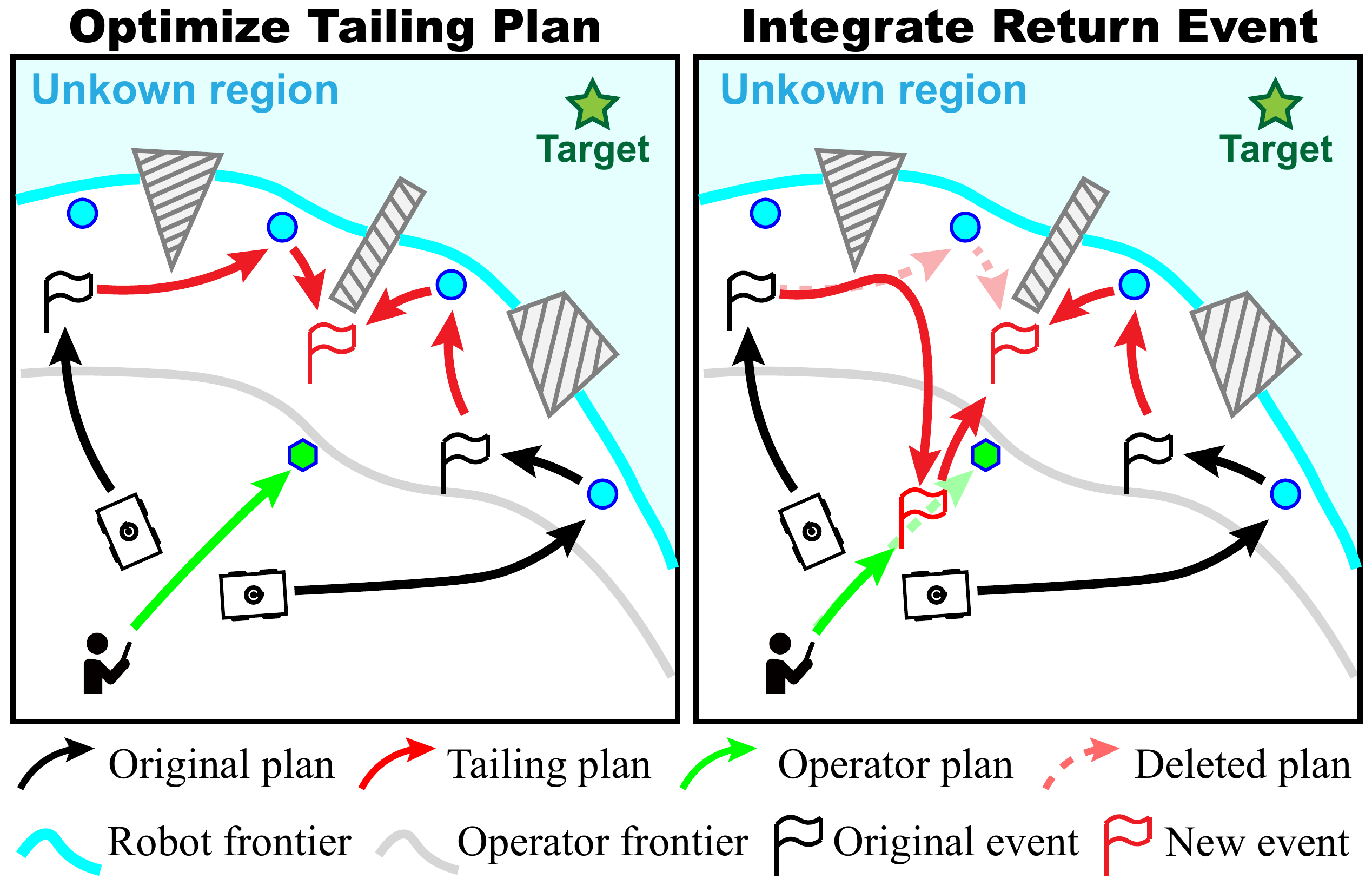}
  \vspace{-0.25in}
  \caption{
    \textbf{Left}:
    When two robots communicate, they first compute their tailing plans,
  including the expoloration paths and the new meeting event.
  \textbf{Right}: return events are then integrated into the tailing plans if needed.
  }\label{fig:Meeting_plan}
  \vspace{-0.1in}
\end{figure}
%==============================

%=======================
\subsubsection{Intermittent Communication with Ring Topology}
\label{subsubsec:ring}
First of all, to facilitate the propagation of information within the team and to the operator,
the robots are designed to meet and communicate intermittently during exploration.
As illustrated in Fig.~\ref{fig:Ring_topo}, the robots start in close proximity,
and they adhere to a fixed ring communication topology.
Specifically, each robot~$i$ is configured to exchange information
exclusively with its predecessor~$i-1$ and its successor~$i+1$, $\forall i = 2, \cdots, N-1$.
Robot~$N$ communicates only with robot~$1$ and robot~$N-1$, while robot~$1$ is connected to robot~$N$ and robot~$2$.
When robot~$i$ and~$j$ communicate at a planned event~$c_{ij}\triangleq (p_{ij}, t_{ij})$, they follow these steps (as illustrated in Fig.~
\ref{fig:Meeting_plan}):
(I) their local maps~$\mathcal{M}_i$ and~$\mathcal{M}_j$ are merged to form a unified map~$\mathcal{M}_{ij}$,
and the observed adversaries are combined into a set~$\mathcal{A}_{ij}\triangleq \mathcal{A}_i \cup \mathcal{A}_j$;
(II) the frontiers~$\mathcal{F}_{ij}$ are extracted from~$\mathcal{M}_{ij}$;
(III) an optimal frontier~$\widehat{f}_i$ is selected for robot~$i$ according to cost function in~\eqref{eq:next-frontier},
where the~$p_\texttt{r}(t)$ is replaced by the next meeting location of robot~$i$;
(IV) similarly, an optimal frontier~$\widehat{f}_j$ is selected for robot~$j$;
(V) a collision free path~$\mathbf{p}_{ij}$ is generated between~$\widehat{f}_i$ and~$\widehat{f}_j$,
within which a meeting point~$p^+_{ij}$ is selected by minimizing the estimated waiting time for both robots, i.e.,
%==============================
\begin{equation}\label{eq:meeting-point}
  p^+_{ij} \triangleq \underset{p \in \mathbf{p}_{ij}}{\textbf{argmin}}
  \,\Big\{{\textbf{max}}\, \{T_i(p),\, T_j(p)\}\Big\},
\end{equation}
%===============================
where $T_i(p)$ and~$T_j(p)$ are the estimated arrival times for robot~$i$ and~$j$ to reach point~$p$;
and the corresponding meeting time is given by~$t^+_{ij} \triangleq \textbf{max}\{T_i(p^+_{ij}),\, T_j(p^+_{ij})\}$.
Therefore, their next event is given by~$c^+_{ij}\triangleq (p^+_{ij}, t^+_{ij})$.
Then, the temporary tailing plan for robot~$i$ is given by
%==============================
\begin{equation}\label{eq:tailing-plan}
  \Gamma^+_i(t_{ij}) = \texttt{HybPlan}(p_{ik},\, \widehat{f}_i,\,
  p^+_{ij},\, \mathcal{M}_{ij}),
\end{equation}
%===============================
where~$p_{ik}$ is the next meeting location of robot~$i$.
The same process is applied to robot~$j$ to obtain its tailing plan~$\Gamma^+_j(t_{ij})$.

%=======================
\subsubsection{Planning of Return Events}
\label{subsubsec:integration-return-event}
In addition to the communication events between robots, they also need to decide whether and when to return to the operator,
as shown in Fig.~\ref{fig:Meeting_plan}.
Assume that robot~$i$ is the predecessor of robot~$j$ in the communication ring,
then they should decide whether robot~$i$ should return to the operator before their next event~$c^+_{ij}$,
since robot~$i$ has the latest information before they meet.
The return event is optimized in the same way as in the optimization of~\eqref{eq:sub-problem},
but with the following modifications:
(I) the robot plan~$\Gamma_\texttt{r}(t)$ is replaced by the tailing plan~$\Gamma^+_i(t_{ij})$;
(II) the starting time~$t$ is replaced by the time of the next communication event of robot~$i$, denoted as~$t_{ik}$.
Resulting solution of the modified problem
is denoted by~$\{p'_i\in \Gamma^+_i(t_{ij}),\, p_\texttt{ret}\in \Gamma_\texttt{h},\, t_\texttt{ret},\, t_\texttt{w}\}$.
If~$p'_i=p^+_{ij}$, it implies that no return event is needed,
and the tailing plan~$\Gamma^+_i(t_{ij})$ remains unchanged;
otherwise, the return event is given by:
\begin{equation}\label{eq:return-event}
  c_{\texttt{ret}}\triangleq (\texttt{h},\, p_\texttt{ret},\, t_\texttt{ret}),
\end{equation}
which is inserted into the tailing plan of robot~$i$
between~$p'_i$ and~$p^+_{ij}$.
Note that the meeting time~$t^+_{ij}$ should be updated accordingly to ensure the feasibility of the plan.
Finally, the complete plan of robot~$i$ at time~$t_{ij}$ is given by~$\widehat{\Gamma}_i\triangleq \Gamma_i+\Gamma^+_i$,
and the same applies to robot~$j$ for its local plan~$\widehat{\Gamma}_j$.

% ==============================
\begin{figure*}[t!]
  \centering
  \includegraphics[width=1.0\linewidth]{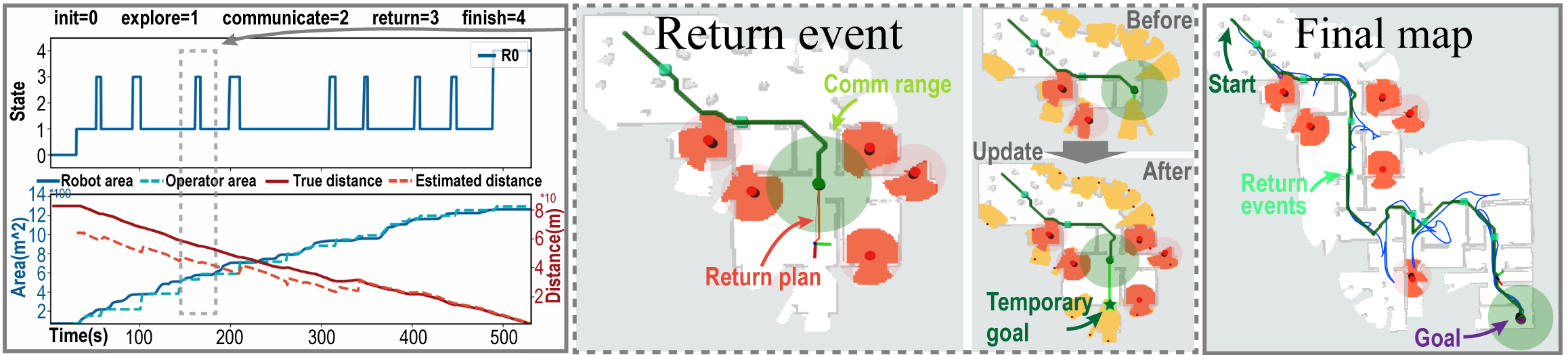}
  \vspace{-0.2in}
  \caption{
  Results of 1 robot escorting the operator in scenario-1 with 6 adversaries (in red).
  \textbf{Left}: Evolution of robot state (explore, communicate and return),
  the area explored by the robot
  and received by the operator,
  and the estimated or remaining distance to the target;
  \textbf{Middle}: One return event by the robot,
  along with the operator map before and after the event;
  \textbf{Right}: Final map and trajectories of the robot (in blue) and the operator (in green).
  }\label{fig:test_sim_1}
  \vspace{-0.1in}
\end{figure*}
%==============================

%==============================
\begin{algorithm}[t!]
\caption{Escorting by $\texttt{BodyGuards}(\cdot)$}
\label{alg:online}
	\LinesNumbered
        \SetKwInOut{Input}{Input}
        \SetKwInOut{Output}{Output}
\Input{$p_\texttt{G}$.}
\Output{$\{\Gamma_i\},\,\Gamma_\texttt{h}$.}
\While{$p_\texttt{h} \neq p_\texttt{G}$}{
\tcc{\textbf{Operator Movement}}
    Update local map~$\overline{\mathcal{M}}_\texttt{h}$ by~\eqref{eq:map-update}\;
    Select temporary goal~$\widehat{p}_\texttt{g}$ by~\eqref{eq:temporary-h-goal}\;
    Generate safe path~$\Gamma_\texttt{h}$ by~\eqref{eq:operator-plan}\;
    Move along~$\Gamma_\texttt{h}$ with speed~$v_\text{h}$\;
  \tcc{\textbf{Multi-robot Coordination}}
  \For{neighbors~$(i,j)$}{
    Compute~$\mathcal{M}_{ij}$,~$\mathcal{A}_{ij}$ and~$\mathcal{F}_{ij}$\;
    Optimize next com. event~$c^+_{ij}$ by~\eqref{eq:meeting-point}\;
    \If{return event required by~\eqref{eq:return-event}}{
        Return to operator and update~$\mathcal{M}_{\texttt{h}}$, $\mathcal{A}_\texttt{h}$\;
    }
  }
  \tcc{\textbf{Online Adaptation}}
    \For{each robot~$i\in \mathcal{N}$}{
        Update local map~$\mathcal{M}_i$\;
        Re-select frontier~$\widehat{f}_i$ by~\eqref{eq:next-frontier}\;
        \If{condition~\eqref{eq:update_frontier} holds}{
            Update local goal to~$\widehat{f}_i$\;
        }
    }
}
\end{algorithm}
% ==============================

%=======================
\subsubsection{Dynamic Adaptation of Exploration Plan}
\label{subsubsec:dynamic-adaptation}
Since the above planning process only occurs at intermittent communication events,
the original local plan may become inefficient due to the newly discovered information during exploration.
In addition, there can be extra time left for exploration before the appointed communication event,
which would be wasted on waiting for the other robot.
Therefore, the robots should dynamically adapt their exploration plans based on their current local map.
Specifically, each time the local map of robot~$i$ is updated, the optimal frontier~$\widehat{f}_i$ is selected by~\eqref{eq:next-frontier}.
Given~$(p_{ik},\, t_{ik})$ as the next meeting event of robot~$i$,
if
\begin{equation}\label{eq:update_frontier}
  t+\frac{\chi(p_i,\, \widehat{f}_i,\, p_{ik})}{v_i} \leq t_{ik},
\end{equation}
holds for~$\widehat{f}_i$,
then~$\widehat{f}_i$ is set as the current goal.
Otherwise, the original plan~$\widehat{\Gamma}_i$ is kept.

%%========================================
%%========================================
%===================
\subsection{Overall Framework}
\label{subsec:online}

\subsubsection{Online Execution and Adaptation}
As summarized in Alg.~\ref{alg:online}, the BodyGuards framework integrates operator movement,
multi-robot coordination, and online adaptation in a unified loop. The operator updates the local
map~$\overline{\mathcal{M}}_\texttt{h}$ by~\eqref{eq:map-update}, selects a temporary
goal~$\widehat{p}_\texttt{g}$ by~\eqref{eq:temporary-h-goal}, and generates a safe trajectory
$\Gamma_\texttt{h}$ by~\eqref{eq:operator-plan}. In parallel, robots merge maps and adversary
information to compute joint frontiers, determine the next communication event~$c^+_{ij}$ by
\eqref{eq:meeting-point}, and trigger return events by~\eqref{eq:return-event} if necessary. Each
robot further adapts its plan by re-selecting frontiers with~\eqref{eq:next-frontier} and updating
its goal when the condition in~\eqref{eq:update_frontier} holds.

\subsubsection{Complexity Analysis}
The dominant cost of operator planning comes from frontier search and safe path generation
in \eqref{eq:map-update}--\eqref{eq:operator-plan}, scaling as $\mathcal{O}(n \log n)$ for $n$ map
cells. Robot frontier selection and return optimization
by~\eqref{eq:next-frontier},~\eqref{eq:return-event}) and~\eqref{eq:update_frontier} are bounded by
$\mathcal{O}(k n \log n)$ for $k$ candidate frontiers. Multi-robot coordination, including map
merging and meeting-point optimization~\eqref{eq:meeting-point}, grows as $\mathcal{O}(Nn)$ per cycle
for $N$ robots. These bounds confirm real-time feasibility of the framework under typical scenarios.

%%========================================

%%========================================
%==============================
\section{Numerical Experiments} \label{sec:experiments}

For further validation,
numerical simulations and hardware experiments are presented in this section.
The proposed method is implemented in \texttt{Python3}
within the framework of \texttt{ROS},
and tested on a computer with an Intel Core i7-13700KF CPU.
Simulation and experiment videos can be found in the supplementary files.

%====================
\subsection{System Setup}\label{subsec:system}
The robotic fleet consists of~$4$ differential-driven UGVs,
which are  simulated in the \texttt{Stage} simulator and
visualized in the \texttt{Rviz} interface.
As shown in Fig.~\ref{fig:test_sim_1} and~\ref{fig:compare_fig}, two different workspaces are tested:
(I) a large office building surrounded by the forest of size~$50m\times 50m$;
and (II) a large subterranean cave of size~$65m\times 57m$ with numerous tunnels.
The occupancy grid map~\cite{moravec1985high} is adopted with a resolution of~$0.2m$,
and generated via the SLAM package~$\texttt{gmapping}$.
Each robot navigates using the navigation stack \texttt{move\_base},
with a sensor range of~$8m$,
a maximum linear velocity of~$0.8m/s$ and angular velocity of~$1.5rad/s$.
The operator could move with a velocity of~$0.3m/s$.
%==============================
Moreover, two robots can only communicate if they are within a range of~$5m$ and have a line-of-sight,
the same between robots and the operator.
Merging of local maps during communication are handled by the ROS package~$\texttt{multirobot\_map\_merge}$.
As shown in Fig.~\ref{fig:test_sim_1} and~\ref{fig:simulation_1_m},
numerous adversaries are scattered within the environment and initially unknown,
and are discovered by the robots within the range of~$3m$.

%====================
\subsection{Simulation Results}\label{subsec:simulation_results}
To begin with, the proposed BodyGuards framework is tested in scenario-1 with~$1$ robot and $6$ adversaries,
as shown in Fig.~\ref{fig:test_sim_1}.
The robots and the operator start from the top left corner and need to reach the target at the bottom right corner.
During execution, each planning process takes around~$0.2s$ for the robot, which enables an exploration efficiency of $2.4m^2/s$,
and results in~$9$ return events to the operator.
This enables the operator to receive a total area of~$1300m^2$ of explored space.
With the received information, the operator replans its path to the target every~$2s$,
and avoids the potential risk zones estimated from the propagated adversaries.
Finally, the operator reaches the target safely at~$532s$ after moving~$90m$,
and avoids all the adversaries successfully with a minimal distance of~$3.2m$,
which is greater than the safety distance of~$3m$.
Note that only around~$52\%$ of the environment is explored when the operator reaches the target,
indicating the efficiency of the proposed method in completing the task without full exploration.

\begin{figure}[t!]
  \centering
  \includegraphics[width=1.0\linewidth]{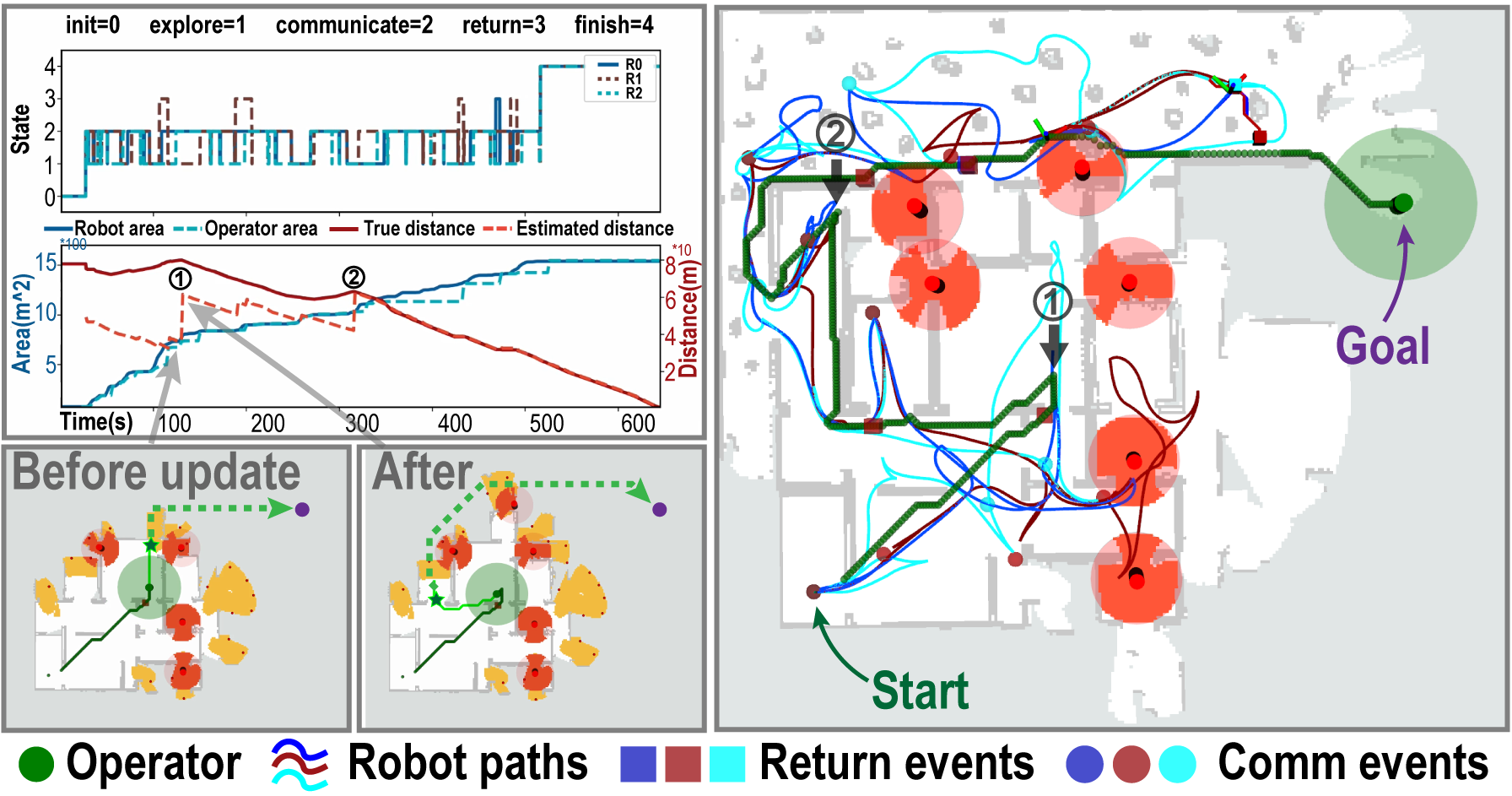}
  \vspace{-0.25in}
  \caption{
  Simulation results of 3 robots escorting the operator in scenario-1 with 6 adversaries.
  Since the optimal path is blocked by adversaries, a drastic increase in estimated distance occurs
  when the adversaries are discovered.
  }\label{fig:simulation_1_m}
  \vspace{-0.1in}
\end{figure}

Furthermore, a team of 3 UGVs is tested in the same scenario with a more challenging setting,
as shown in Fig.~\ref{fig:simulation_1_m}.
Different from the single-robot case, the three robots change their modes dynamically between exploration, communication and returning,
with in total~$21$ meeting events and~$6$ return events, where each meeting event takes around~$0.3s$ to plan.
With the cooperative behavior, the three robots can explore with a higher efficiency of~$3.5m^2/s$,
almost~$50\%$ higher than the single-robot case.
However, the optimal path to the target is blocked by the adversaries in this case,
resulting in a drastic increase in estimated distance from~$30m$ to~$60m$ when the adversaries are discovered
and propagated to the operator at~$132s$.
This information enables the operator to replan his path in time and change to the correct direction.
Finally, the operator reaches the target safely at~$643s$ after moving~$118m$.
This task is completed after exploring~$1500m^2$ of the environment,
which is around~$60\%$ of the whole area.

% ==============================
%==============================

% ==============================
\begin{figure}[t]
  \centering
  \includegraphics[width=1.0\linewidth]{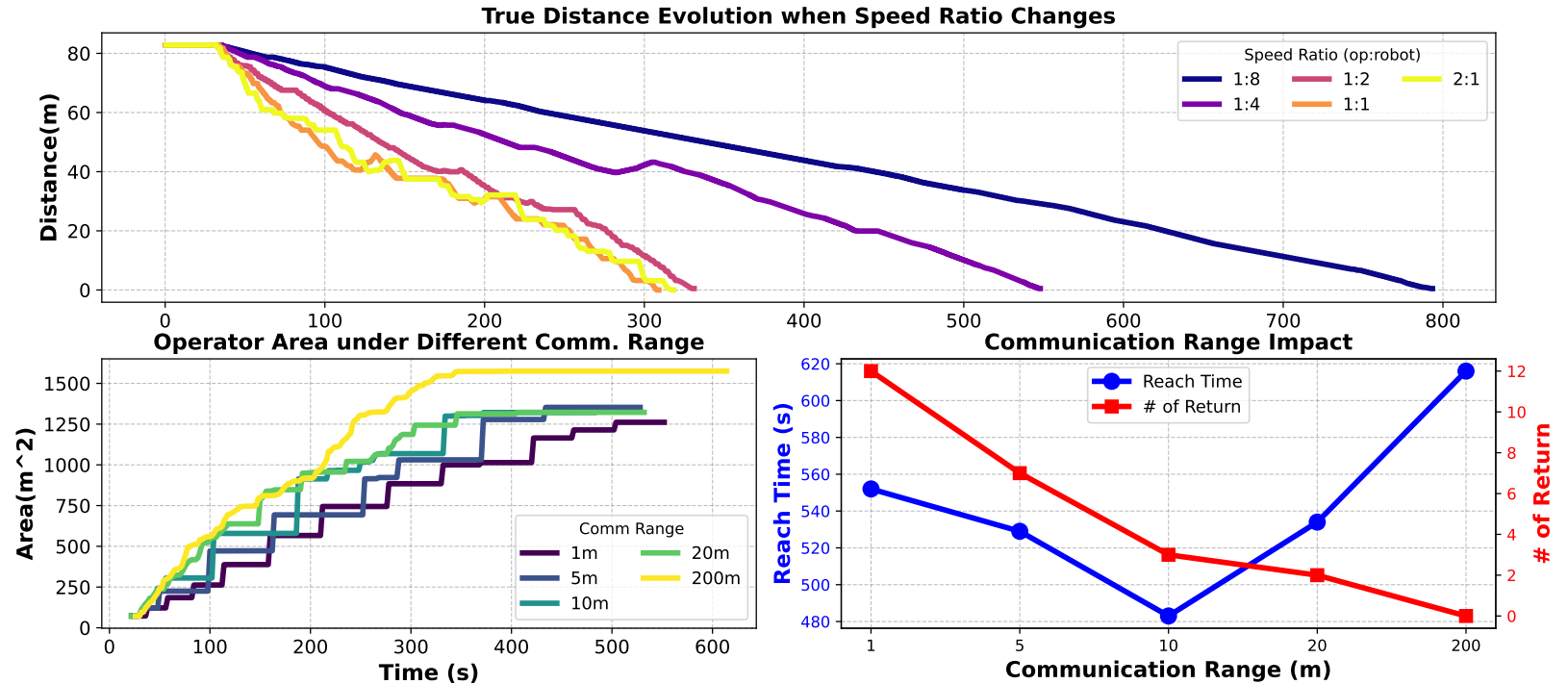}
  \vspace{-0.25in}
  \caption{
    Sensitivity analyses of the proposed dual-mode exploration and communication
    strategy under different operator-robot velocity ratios (\textbf{Top})
    and different communication ranges (\textbf{Bottom}).
  }\label{fig:sensitivity}
  \vspace{-0.1in}
\end{figure}
%==============================

%====================
\subsection{Sensitivity Analyses}\label{subsec:parameter}
Sensitivity of the proposed method is further analyzed w.r.t.
\emph{two key parameters}:
(I) the velocity ratio between the operator and robots;
and (II) the communication range.
As summarized in Fig.~\ref{fig:sensitivity},
when the robot velocity is fixed at $0.8\,\text{m/s}$ and the operator velocity increases,
a nearly proportional reduction in the mission time can be observed.
However, when the velocity ratio exceeds $0.5$, further increasing the operator velocity provides little improvement in overall efficiency.
In this regime, the operator can almost maintain real-time communication with the robot,
and the task performance becomes primarily constrained by the robot's velocity.
On the other hand, as the communication range increases from~$1m$ to~$200m$,
the operator receives more map area from $1250m^2$ to~$1500m^2$ with less return events from $12$ to $0$.
However, the overall task time can be maintained below~$650s$ in all cases,
and even~$550s$ when the communication range is limited to only~$1m$.
This demonstrates the robustness of the proposed method under different communication constraints.

%==============================
\begin{table}[t!]
\centering
\caption{Comparison of Baseline Methods (Avg. 5 Runs).}\label{tab:comparison}
\small   % 调大字体
\renewcommand{\arraystretch}{1.0}  % 行距稍微放宽
\setlength{\tabcolsep}{3pt}        % 列间距加宽一点以占满一栏
\begin{tabular}{c|c|cc|cc|cc}
\toprule
\multirow{2}{*}{\textbf{Scenario}} & \multirow{2}{*}{\textbf{Method}}
& \multicolumn{2}{c|}{\textbf{Find T. (s)}}
& \multicolumn{2}{c|}{\textbf{Reach T. (s)}}
& \multicolumn{2}{c}{\textbf{Op. Dist. (m)}} \\
\cmidrule(lr){3-4} \cmidrule(lr){5-6} \cmidrule(lr){7-8}
 & & Case1 & Case2 & Case1 & Case2 & Case1 & Case2 \\
\midrule
\multirow{6}{*}{\textbf{Office}}
 & \textbf{Ours} & \textbf{468} & 600 & \textbf{528} & \textbf{682} & \textbf{91} & \textbf{118} \\
 & TARE & 748 & 754 & 784 & 874 & 116 & 143 \\
 & POI  & 507 & \textbf{580} & 600 & 687 & 103 & 119 \\
 & FSMP & 578 & 819 & 641 & 838 & 101 & 139 \\
 & ETC  & 534 & 749 & 641 & 845 & 93 & 169 \\
 & TTC  & 516 & 640 & 580 & 873 & 93 & 157 \\
 & N-PRZ & -- & --  & --  & --  & -- & -- \\
\midrule
\multirow{6}{*}{\textbf{Cave}}
 & \textbf{Ours} & \textbf{653} & \textbf{1025} & \textbf{731} & \textbf{1094} & \textbf{139} & 202 \\
 & TARE & 887 & 1232 & 953 & 1312 & 153 & 260 \\
 & POI  & 848 & 1050 & 933 & 1115 & 167 & \textbf{194} \\
 & FSMP & 918 & 1060 & 996 & 1148 & 161 & 219 \\
 & ETC  & 724 & 1120 & 844 & 1220 & 156 & 241 \\
 & TTC  & 770 & 1130 & 835 & 1225 & 145 & 206 \\
 & N-PRZ & -- & --  & --  & --  & -- & -- \\
\bottomrule
\end{tabular}
\end{table}
%==============================

%====================
\subsection{Comparisons}\label{subsec:comparison}
\begin{figure}[t!]
  \centering
  \includegraphics[width=1.0\linewidth]{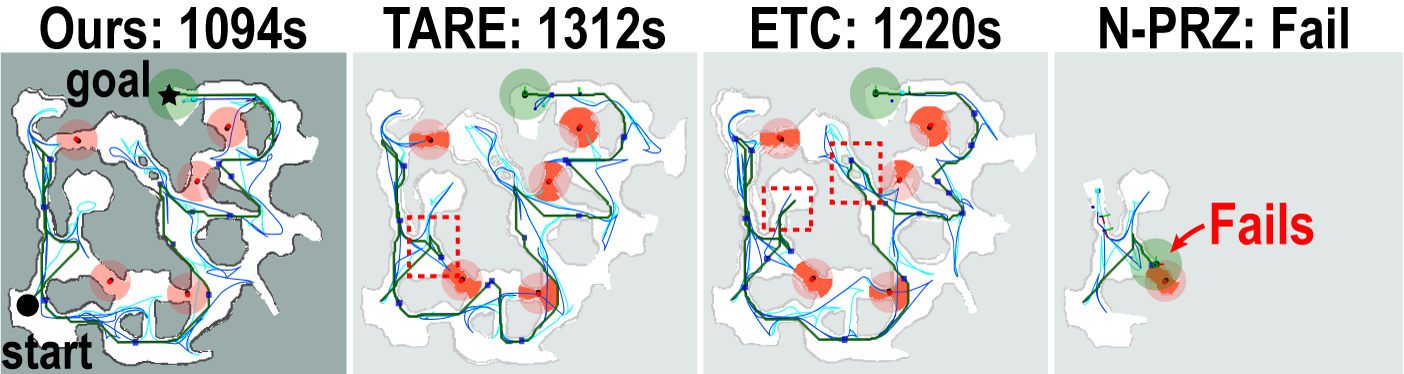}
  \vspace{-0.25in}
  \caption{
    Simulation results of the baseline methods in the Scenario-2,
    where the red squares highlight the detour caused by inefficient exploration.
  }\label{fig:compare_fig}
\vspace{-0.1in}
\end{figure}

The proposed framework (\textbf{Ours}) is compared with~$6$ baselines:
(I) \textbf{TARE}, a exploration framework which does not consider the target information~\cite{cao2023tare};
(II) \textbf{POI}, which generates points of interet to guide exploration~\cite{9645287};
(III) \textbf{FSMP}, a sampling-based planner for robotic exploration~\cite{10909355};
(IV) \textbf{ETC}, an event-triggerd communication policy where the robots return only when the operator reaches temporal goal;
(V) \textbf{TTC}, a time-triggerd return policy where return event occurs every~fixed period;
(VI) \textbf{N-PRZ}, our method without considering the potential risk zones.
The compared metrics are the time that the robotic fleet find the target, the time for the oprtrator to reach target location,
and the total distance that the operator travels.
Each method is tested~$5$ times in both scenarios with single-robot as Case1 and two-robot as Case2.
The results are summarized in Table.~\ref{tab:comparison} and Fig.~\ref{fig:compare_fig}.

% ==============================
%==============================

It can be seen that the proposed framework consistently outperforms the baselines across all evaluation metrics in both scenarios.
Methods such as \textbf{TARE} and \textbf{FSMP}, which do not explicitly consider goal-reaching objectives,
exhibit longer finding and reaching times (e.g., \textbf{TARE} 748s vs. \textbf{Ours} 468s in office Case 1).
A similiar phenomenon can be observed for \textbf{POI}, which takes around~$25\%$ more time in finding target than our method in all cases.
On the other hand, \textbf{ETC} and \textbf{TTC} present much larger reaching time and longer operator trajectory owing to non-optimal return events.
It also results in more detour in the operator path, as shown in Fig.~\ref{fig:compare_fig}.
Lastly,~\textbf{N-PRZ} presents a success rate of $0\%$ in all tests,
demonstrating the necessity of considering the PRZ regions during planning.

%==========================================
\subsection{Hardware Experiments}\label{subsec:hardware}

% ==============================
\begin{figure}[t!]
  \centering
  \includegraphics[width=1.0\linewidth]{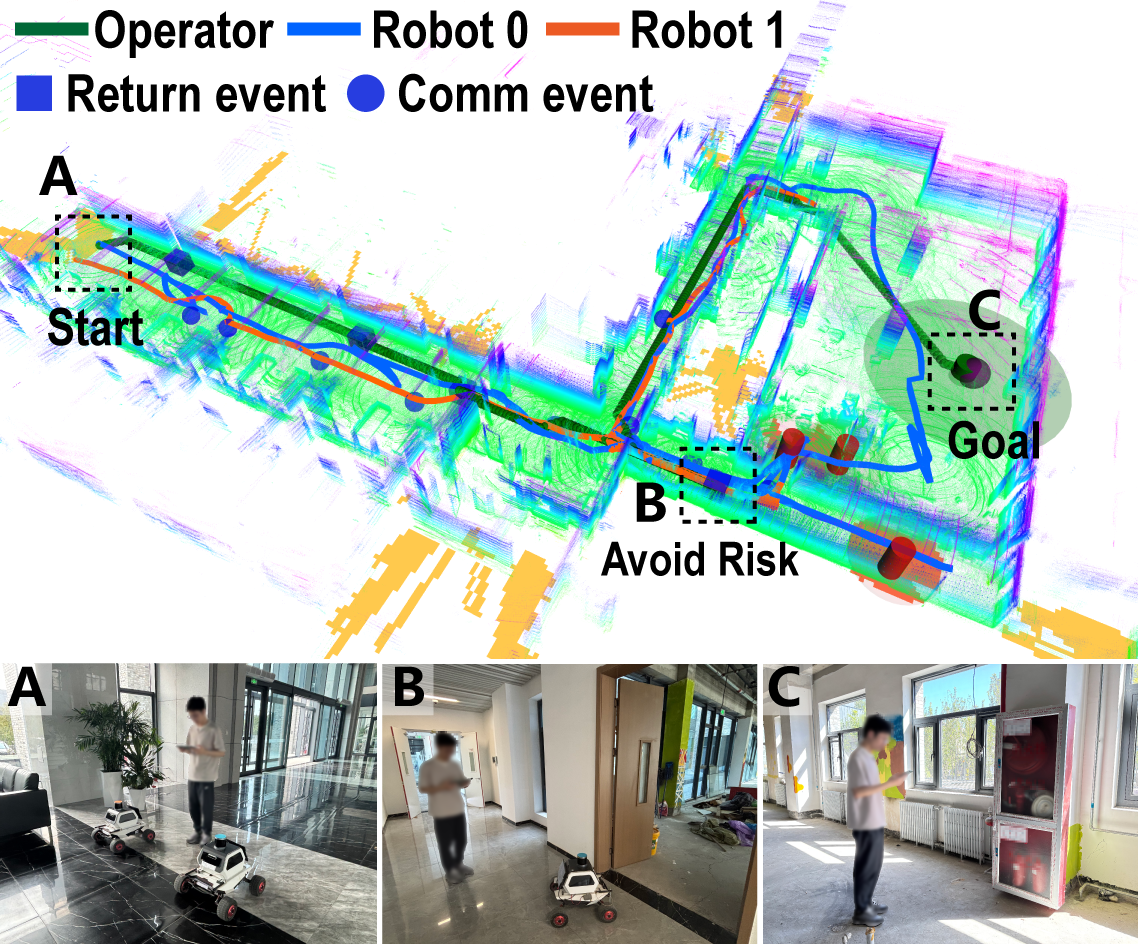}
  \vspace{-0.25in}
  \caption{
    The pointcloud map and trajectories from hardware experiments (\textbf{Top}),
    and snapshots during the experiments (\textbf{Bottom}).
  }\label{fig:real_exp}
\vspace{-0.1in}
\end{figure}
%==============================

As shown in Fig.~\ref{fig:overall} and~\ref{fig:real_exp}, an operator deploys 2 Scout-mini UGVs
for escorting in an office environment of size~$60m \times 40m$,
and interacts with the fleet with a tablet.
Each robot and the operator are equipped with an ad-hoc network device for
close-range communication (AP-DLINK1402A).
They all start from the left corner of the corridor in Fig.~\ref{fig:real_exp}-A,
and aims to escort the operator to the extinguisher within the room in Fig.~\ref{fig:real_exp}-C.
During the experiment, the two robots plans intotal~$9$ meeting events and~$5$ return events,
and an interesting cooperative behavior of one robot returning while the other robot exploring can be observed in Fig.~\ref{fig:overall}.
As shown in Fig.~\ref{fig:real_exp}-B, 3 risk regions are observed by robot~$0$ at~$450s$,
blocking the optimal path to the target.
Then, this information is relayed to the operator at~$472s$ to enable timely replanning.
Finally, the target point is found by robot~$0$ and known to the operator at~$996s$,
and the operator reaches the target safely at~$1350s$ after moving~$94m$.
It is worth noting that the task is completed without full exploration of the environment ($80\%$ coverage),
which differs our method from traditional strategies that prioritize complete mapping
and demonstrates the efficiency of the proposed method.

%%========================================
\section{Conclusion} \label{sec:conclusion}
This work proposes a novel and generic framework as \textbf{BodyGuards}
for the online escorting task by the robotic fleet
in unknown and communication-constrained environments.
Future work involves combining map prediction techniques and semantic information into planning.

%%========================================

% \newpage
%========================================
\bibliographystyle{IEEEtran}
\bibliography{contents/references}

\end{document}